\PassOptionsToPackage{many}{tcolorbox} 
\documentclass[11pt]{article}

\usepackage[final]{acl}

\usepackage{times}
\usepackage{latexsym}
\usepackage[utf8]{inputenc}
\usepackage{amsmath,amssymb}
\usepackage{pifont}

\usepackage{booktabs}
\usepackage{multirow}
\usepackage{lipsum}
\usepackage{tabularx}
\usepackage{subcaption}
\usepackage{arydshln}

\usepackage{makecell}

\usepackage{siunitx}
\usepackage{hyperref}
\hypersetup{
    colorlinks=true,
    linkcolor=black,
    filecolor=magenta,      
    urlcolor=magenta, 
    pdftitle={VideoDR Benchmark},
}

\usepackage[T1]{fontenc}

\usepackage[utf8]{inputenc}

\usepackage{microtype}

\usepackage{inconsolata}

\usepackage{graphicx}
\usepackage{enumitem}
\usepackage{pifont}

\usepackage{tcolorbox}

\title{\hspace{-1.5em}\raisebox{-0.3\height}{\includegraphics[height=1.9em]{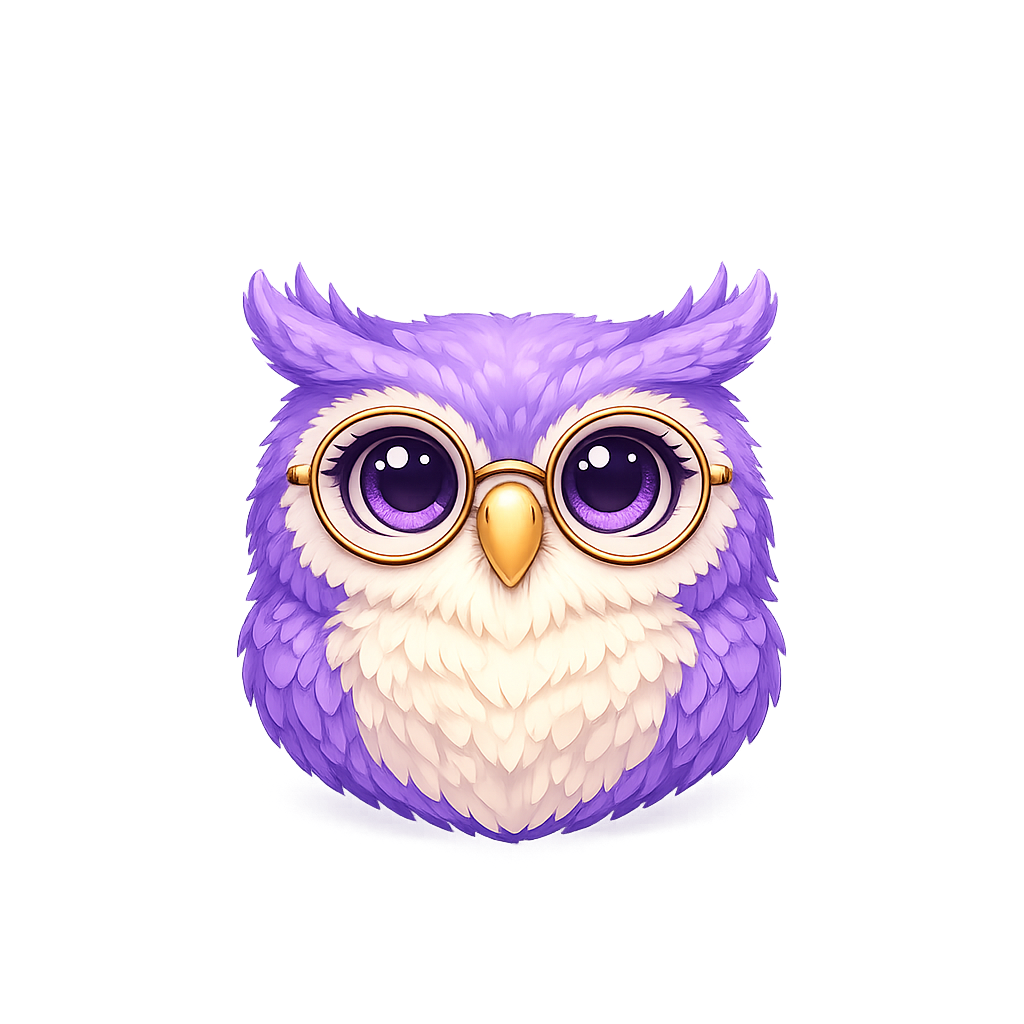}} Watching, Reasoning and Searching: A Video Deep Research Benchmark\\on Open Web for Agentic Video Reasoning}

\author{
    \textbf{Chengwen Liu\textsuperscript{1*}},
    \textbf{Xiaomin Yu\textsuperscript{2*}},
    \textbf{Zhuoyue Chang\textsuperscript{1*}},
    \textbf{Zhe Huang\textsuperscript{10*}},
    \textbf{Shuo Zhang\textsuperscript{10*}},
    \textbf{Heng Lian\textsuperscript{10}},
    \\
    \textbf{Jisheng Dang\textsuperscript{1\dag}},
    \textbf{Rui Xu\textsuperscript{4}},
    \textbf{Sen Hu\textsuperscript{5,10}},
    \textbf{Jianheng Hou\textsuperscript{6}},
    \textbf{Chengwei Qin\textsuperscript{2,9}},
    \textbf{Xiaobin Hu\textsuperscript{7}},
    \textbf{Kunyi Wang\textsuperscript{3,10}},
    \\
    \textbf{Zhi Yang\textsuperscript{10}},
    \textbf{Hao Peng\textsuperscript{1}},
    \textbf{Hong Peng\textsuperscript{1\dag}},
    \textbf{Ronghao Chen\textsuperscript{5,10\dag}},
    \textbf{Huacan Wang\textsuperscript{8,10\dag}}
    \\
    \textsuperscript{1}LZU,
    \textsuperscript{2}HKUST(GZ),
    \textsuperscript{3}UBC,
    \textsuperscript{4}FDU,
    \textsuperscript{5}PKU,
    \textsuperscript{6}USC,
    \textsuperscript{7}NUS,
    \textsuperscript{8}UCAS,
    \textsuperscript{9}HKUST,
    \textsuperscript{10}QuantaAlpha,
    \\
    \small{\textbf{\textsuperscript{*}These authors contributed equally to this work.}}
    \\
    \small{
        \textbf{Project Leader:} \textbf{Xiaomin Yu} \href{mailto:yuxm02@gmail.com}{yuxm02@gmail.com}
    }
    \\
    \small{
        \textbf{\dag Correspondence:} \href{mailto:pengh@lzu.edu.cn}{pengh@lzu.edu.cn},
        \href{mailto:dangjisheng@lzu.edu.cn}{dangjisheng@lzu.edu.cn},
        \href{mailto:chenronghao@alumni.pku.edu.cn}{chenronghao@alumni.pku.edu.cn},
        \href{mailto:wanghuacan17@mails.ucas.ac.cn}{wanghuacan17@mails.ucas.ac.cn}
    }
    \\
    \small{
        \textbf{Github:} \url{https://github.com/QuantaAlpha/VideoDR-Benchmark}
    }
}

\begin{document}
\maketitle
\def\method{VideoDR}

\begin{abstract}
In real-world video question answering scenarios, videos often provide only localized visual cues, while verifiable answers are distributed across the open web; models therefore need to jointly perform cross-frame clue extraction, iterative retrieval, and multi-hop reasoning-based verification. To bridge this gap, we construct the first video deep research benchmark, \method{}. \method{} centers on video-conditioned open-domain video question answering, requiring cross-frame visual anchor extraction, interactive web retrieval, and multi-hop reasoning over joint video–web evidence; through rigorous human annotation and quality control, we obtain high-quality video deep research samples spanning six semantic domains. We evaluate multiple closed-source and open-source multimodal large language models under both the Workflow and Agentic paradigms, and the results show that Agentic is not consistently superior to Workflow: its gains depend on a model’s ability to maintain the initial video anchors over long retrieval chains. Further analysis indicates that goal drift and long-horizon consistency are the core bottlenecks. In sum, \method{} provides a systematic benchmark for studying video agents in open-web settings and reveals the key challenges for next-generation video deep research agents.
\end{abstract}

\begin{figure*}[h]
  \centering
  \includegraphics[width=\linewidth]{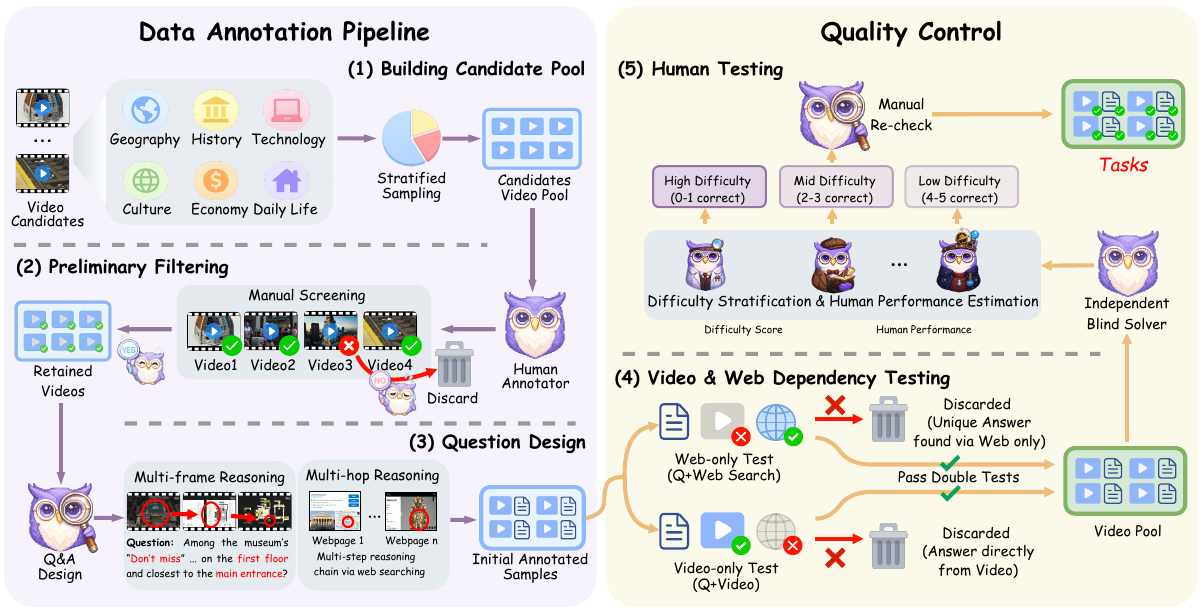}
  \caption{Overview of the \method{} construction pipeline.}
  \label{fig:vdr_pipeline}
\end{figure*}

\newpage

\section{Introduction}

In existing multimodal evaluations, video remains a significant weakness \cite{li2024mvbench,fang2024mmbench,wu2024longvideobench,jang2024videowebarena}.
On the one hand, video reasoning inherently requires cross-temporal cue tracking and spatiotemporal modeling \cite{wu2024longvideobench,yang2025longvt,chen2025lvagent};
on the other hand, most existing evaluations adopt a closed-evidence setting, where models typically only need to answer questions within the given video, without interacting with evidence on the open web 
\cite{fu2025video,zhou2025mlvu,wang2025lvbench,yang2025svbench}.
As a result, the capability of using videos as clues and completing fact verification and reasoning synthesis on open webpages has not been systematically characterized.

Meanwhile, deep research agents are pushing question answering from static contexts toward active evidence exploration on the open web: instead of answering directly from a given context, systems must conduct multiple rounds of search, filtering, and cross-checking in real web environments, ultimately producing conclusions grounded in evidence \cite{chen2025learning,zheng2025deepresearcher}.
A large number of deep research benchmarks have emerged around this capability \cite{li2025search,jin2025search}.
However, overall, these benchmarks still mostly start from textual queries \cite{wei2025browsecomp,wu2025webwalker,li2025webthinker}; even when multimodal information is introduced, visual content is often treated as static auxiliary information rather than key evidence that must be precisely tracked and propagated \cite{jiang2024mmsearch}. 


However, in real use, videos often carry decisive clues. User questions about videos are typically open-domain factoid questions: the knowledge to be verified does not directly appear in the video or its title, but is distributed across large and dynamically changing web corpora; meanwhile, the key clues relevant to the question lie along the video timeline and must be extracted through cross-frame association. In this research pattern, videos provide localized visual cues, and webpages provide verifiable answers—yet it is not directly covered by existing deep research benchmarks that take text as input \cite{wei2025browsecomp}, nor by video benchmarks that assume evidence is closed within the video \cite{li2024mvbench,fang2024mmbench,zhou2025mlvu}.

\begin{figure}[t]
  \centering
  \includegraphics[width=0.8\linewidth]{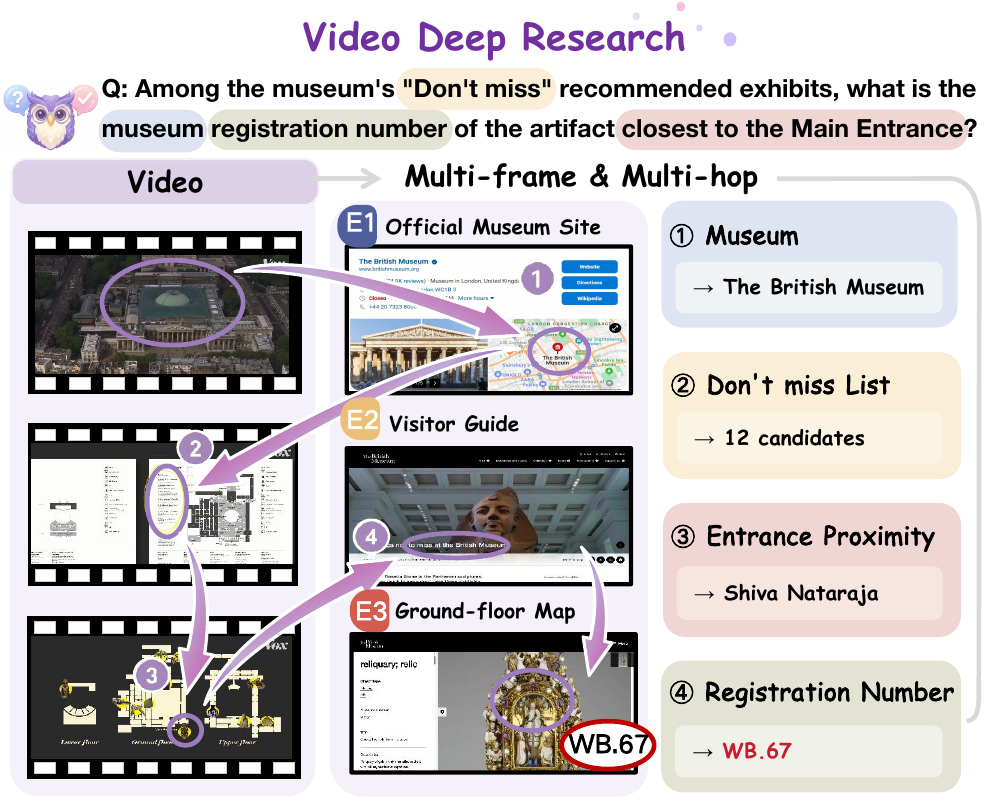}
  \caption{An example of the \method{} task: identifying a museum via video visual cues, then using multi-hop search to find the closest "don't miss" exhibit to the entrance and outputting its accession number WB.67.}
  
  \label{fig:vdr_case}
\end{figure}

Based on these gaps, we propose \method{}, the first open-domain benchmark that systematically evaluates video deep research.
As shown in \autoref{fig:vdr_case}, we extend deep research to an open-domain factoid question answering setting conditioned on video: models must extract and compose visual anchors from multiple frames \cite{yang2025longvt,chen2025lvagent,zhang2025deep},
use browser-based search to locate candidate evidence on the open web \cite{wei2025browsecomp,wu2025webwalker,li2025webthinker,chen2025learning,zheng2025deepresearcher},
and perform multi-hop reasoning in the joint evidence space of videos and webpages to output a unique and verifiable answer \cite{liang2025video}.
To support this setting, we explicitly incorporate an interactive web search process into the task definition and adopt annotation and strict quality control over diverse real-world scenarios to systematically remove samples that can be answered by the video alone or by webpages alone, making the combined capability of video understanding, web search, and evidence-based reasoning the core evaluation target \cite{yao2022react,gao2023retrieval}.
Based on VideoDR, we evaluate mainstream models under both the Workflow and Agentic paradigms~\citep{liu2025advances}.
The evaluated models include closed-source models Gemini-3.1-pro-preview~\citep{google2026gemini31propreview} and GPT-5.4~\citep{openai2026gpt54}, as well as open-source models Qwen3.5-35B-A3B~\citep{qwen2026qwen35}, InternVL3.5-14B~\citep{wang2025internvl35}, Gemma-4-31B-it~\citep{googledeepmind2026gemma4}, and GLM-4.6V-Flash~\citep{zai2025glm46vflash}.
We conduct comprehensive performance and error analyses along three dimensions: difficulty, video duration, and semantic domain.

\paragraph{Our main contributions are as follows:}

\begin{itemize}[leftmargin=15pt]
    \item[\ding{182}] \textbf{Video Deep Research Task}: We first define the Video Deep Research task, shifting video understanding from closed-context perception to active, multi-hop search and reasoning on the open web anchored by video cues.
    
    \item[\ding{183}] \textbf{\method{} Benchmark}: We construct a high-quality \method{} benchmark through rigorous human annotation and quality control. \method{} benchmark ensures that the multi-step evidence gathering process maintains a strong dependency on multi-frame visual cues within the video.
 
    \item[\ding{184}] \textbf{Agent Capability Boundaries}: By benchmarking leading MLLMs across Workflow and Agentic paradigms, we delimit the capability boundaries of these two agentic approaches. Leveraging the diverse distribution of VideoDR across semantic domains, question lengths, and video durations, we systematically analyze the performance and error patterns under different paradigms. Our findings reveal that Goal Drift and Long-horizon Consistency are the core bottlenecks constraining the development of next-generation video deep research agents.

\end{itemize}


\section{Related Work}

\textbf{Deep Research Benchmarks.} Existing deep research evaluations typically assess search, reasoning, and tool use as a unified process \cite{wei2025browsecomp,wu2025webwalker,li2025webthinker,zheng2025deepresearcher}: one line of work tests multi-step query planning and information localization in live web environments \cite{wei2025browsecomp,wu2025webwalker}, while another studies more controllable settings to improve reliability and stability \cite{xue2025simpletir,chen2025learning,jin2025search,li2025search}. Overall, however, most benchmarks still start from textual queries \cite{wei2025browsecomp,wu2025webwalker}, and visual content is often downplayed as static auxiliary information rather than first-class evidence in the retrieval and verification loop \cite{jiang2024mmsearch,liang2025video,liu2025advances}.

\textbf{Video Reasoning Benchmarks.} Existing video reasoning evaluations are conducted under a closed-evidence setting, where questions are designed to be answerable using only the video itself, and they primarily stress long-video temporal understanding and long-context reasoning \cite{fu2025video,li2024mvbench,wu2024longvideobench,zhou2025mlvu,wang2025lvbench,fang2024mmbench,yang2025svbench,li2024videovista,nagrani2024neptune,yu2025vrbench}.
Recent agentic video efforts also explore multi-round interaction and tool calling within the video-understanding loop \cite{zhang2025deep,yang2025longvt,chen2025lvagent}. In summary, systematic evaluations of multi-step evidence gathering and reasoning integration on the open web anchored to video cues remain relatively scarce.

\section{Video Deep Research}

\subsection{Task Definition}

We propose \textbf{Video Deep Research (\method{})}: an open-domain factoid question answering benchmark conditioned on a given video, designed to evaluate a model's ability to perform complex reasoning anchored in the video while leveraging the open web. Given a video $V$ and a natural language question $Q$, the model can interactively call a browser search tool $S$, iteratively searching between video cues and web-page evidence, and finally output a factual answer $A$. During the subsequent research process, the model is not allowed to repeatedly re-watch the video. Each \method{} sample is constructed such that the model must exploit multi-frame cues from the video to locate candidate evidence on the open web, and then perform multi-hop reasoning over the joint evidence space of the video and the open web in order to obtain a unique answer. Formally, the \method{} task can be expressed as:
\[
f: (V, Q; S) \rightarrow A.
\]

\subsection{Data Annotation Process}

In the data construction stage, we recruited three annotators with experience in video understanding and web search to create questions and annotate answers. Each annotator must actively locate several multi-frame visual cues in the video and design corresponding multi-hop questions and answers around these cues. To ensure annotation consistency, we provided unified annotation guidelines and examples before the annotation phase. As shown in \autoref{fig:vdr_pipeline}, the overall annotation pipeline consists of three steps.

\paragraph{Candidate Video Pool.} Annotators first select videos from different platforms and then perform stratified sampling along three dimensions: source, domain, and duration, to cover diverse real-world scenarios. To ensure data quality, we apply a strict negative filtering strategy to remove the following three types of videos: \ding{182} single-scene clips with highly redundant visual semantics; \ding{183} popular topics whose information is overly explicit and can be obtained via text search without watching the video; \ding{184} isolated content on the open web for which no verifiable chain of evidence can be found.

\paragraph{Initial Filtering.} At the initial screening stage, we manually remove clips that lack prominent \emph{visual anchors}. We only retain videos that exhibit coherent visual cues at multiple time points and are suitable for cross-frame association, as candidates for subsequent annotation. For longer videos, annotators are allowed to extract multiple semantically focused segments from the same video and construct separate questions for each segment.

\paragraph{Question Design.} As shown in \autoref{fig:vdr_case}, in this stage, annotators are required to design high-complexity questions for the retained video segments, under two strict constraints: \textbf{\ding{182} Multi-frame reasoning:} Each question must be grounded in video cues spanning multiple frames; it should be impossible to answer the question from a single-frame screenshot. \textbf{\ding{183} Multi-hop reasoning:} Each question must implicitly admit a decomposable multi-step reasoning path, forcing the model to perform at least one round of information exchange between video perception and external search. To ensure verifiability, we archived the web pages containing the key evidence supporting each answer.

\subsection{Quality Control}

To ensure that \method{} is rigorous and unambiguous, we apply a two-stage quality verification procedure to the annotated samples.

\subsubsection{Video \& Web Dependency Testing}

To verify that the annotated samples simultaneously depend on both the video and search, we design two ablation settings and conduct tests on all samples. Only samples that fail under both of the following conditions are retained as valid:
\paragraph{\textbf{Web-only Test.}} The annotator is only given the textual question $Q$ and can conduct the web search. If a unique and unambiguous answer can be obtained purely through web search, the question is deemed to lack dependence on visual anchors and is discarded.
\paragraph{\textbf{Video-only Test.}} The annotator answers the question by watching only the video $V$. If the answer can be directly obtained from the video information alone, the question degenerates into a pure video understanding task and is discarded.

\begin{figure}[t]
  \centering
  \includegraphics[width=0.88\linewidth]{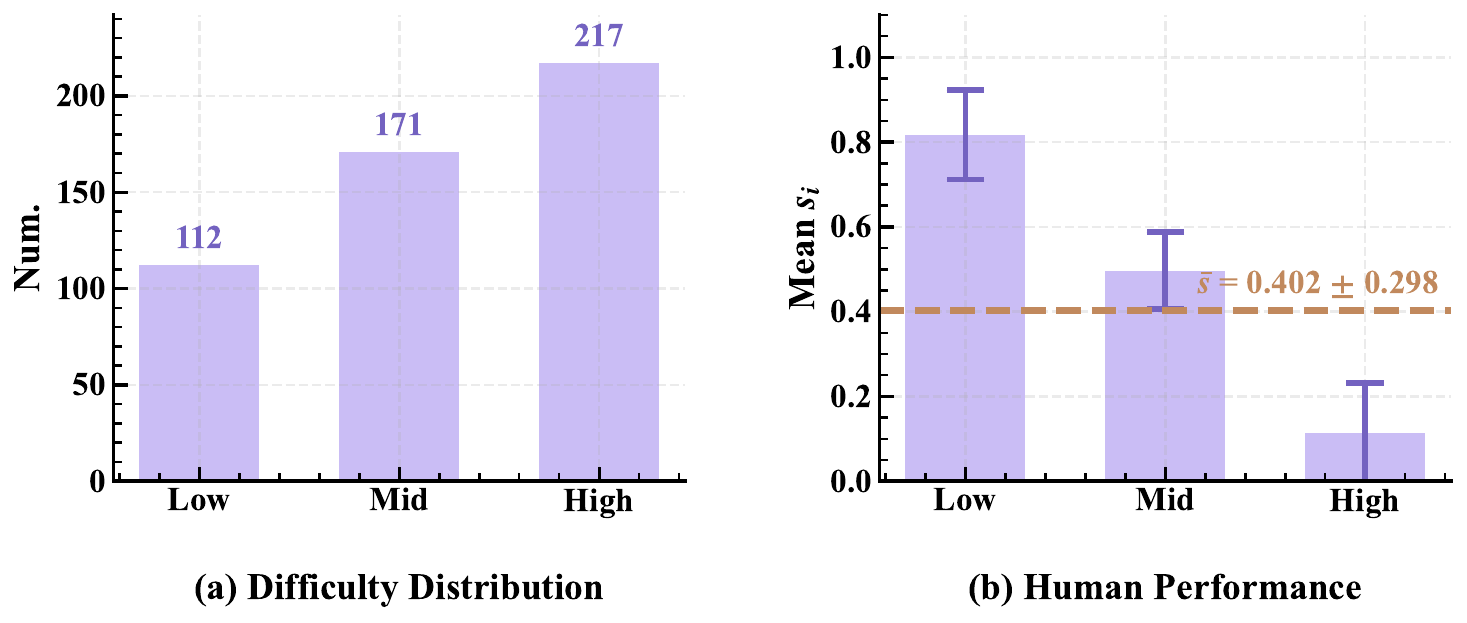}
  \caption{Human solvability across benchmark difficulty levels.}
  \label{fig:vdr_human}
\end{figure}

\subsubsection{Human Testing}

To verify the correctness of annotated questions and characterize their difficulty, we adopt a human evaluation protocol with five independent participants solving the tasks in a blind manner. For each sample $i$, five subjects independently produce answers with access to the annotated video, the textual question, and autonomous web search, but without being given any reference answers, forming the answer set $\{a_{i1}, \ldots, a_{i5}\}$. Subjects are required to complete the task in a browser environment and must, for each question, submit a final answer obtained via video browsing and autonomous web search. This process serves two purposes: \ding{182} to validate the solvability of the questions and the correctness of the annotations under conditions close to real-world usage; and \ding{183} to provide empirical evidence for subsequent difficulty stratification and estimation of the human reference performance.

\paragraph{Difficulty.}
To quantify question difficulty at the sample level, we define the difficulty score of sample $i$ based on the human success rate from the five-subject blind evaluation:
\[
s_i = \frac{1}{5}\sum_{j=1}^{5}\mathbf{1}[a_{ij} \equiv A_i],
\]
where $a_{ij}$ denotes the answer provided by the $j$-th subject for sample $i$, $A_i$ is the gold answer for that sample, and $\mathbf{1}[\cdot]$ is the indicator function (1 if the prediction is semantically equivalent to the gold answer, and 0 otherwise). We treat the distribution of $\{s_i\}$ as a proxy for question difficulty and stratify samples by the number of correct human answers: if only 0--1 out of 5 subjects answer correctly, the sample is labeled \textit{High}; if 2--3 subjects answer correctly, it is labeled \textit{Mid}; and if 4--5 subjects answer correctly, it is labeled \textit{Low}. This difficulty partition is used in subsequent experiments to analyze how different models perform across human difficulty levels. \autoref{fig:vdr_human}(a) shows the distribution of samples across three difficulty levels.

\paragraph{Human Performance.}
Under the above setting, we estimate the human reference performance by the average accuracy of the 5 subjects over all samples:
\[
\bar{s} = \frac{1}{N}\sum_{i=1}^{N} s_i.
\]
\autoref{fig:vdr_human}(b) shows the human test results. Across all 500 samples, the mean sample-level success rate is $\bar{s}=0.4022$ with a standard deviation of $0.298$. When results are stratified by difficulty, the mean $s_i$ is $0.817\pm0.105$ for Low, $0.496\pm0.091$ for Mid, and $0.114\pm0.117$ for High. Overall, human participants can solve a substantial portion of the samples, and we do not observe a systematic pattern of universal failure, which indirectly supports the consistency of the annotated answers. In addition, during dataset construction, we manually re-examine samples with clear disagreement or samples where multiple mutually inconsistent yet seemingly reasonable answers are provided; if ambiguity remains after multiple rounds of verification, the sample is discarded. The final retained dataset ensures that each question corresponds to a unique and verifiable gold answer, given the available evidence, thereby reducing evaluation noise.

\begin{figure*}[t]
  \centering
  \includegraphics[width=\linewidth]{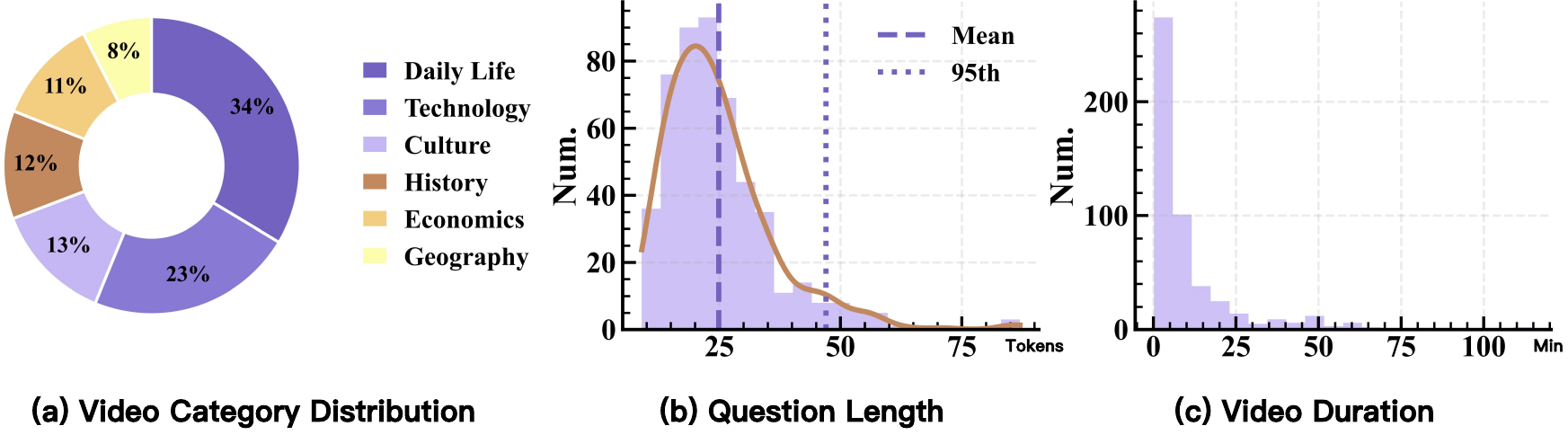}
  \caption{Data statistics of \method{}. (a) video category, (b) question length, and (c) video duration.}
  \label{fig:vdr_stats}
\end{figure*}

\subsection{Data Statistics}


Finally, we construct 500 \method{} samples. To characterize the benchmark’s structural properties, and to provide a basis for stratified analyses in subsequent experiments, we summarize the dataset from three perspectives: \ding{182} video category distribution, \ding{183} question length distribution, and \ding{184} video duration distribution.

\paragraph{\textbf{Video Category Distribution.}} As shown in \autoref{fig:vdr_stats}(a), \method{} spans six semantic domains: Daily Life, Technology, Culture, History, Economics, and Geography. The distribution is relatively balanced: Daily Life accounting for 33.6\%, Technology 22.6\%, Culture 13.0\%, History 11.8\%, Economics 11.4\%, and Geography 7.6\%. This distribution ensures broad coverage of diverse content types in open-domain settings.

\paragraph{\textbf{Question Length.}}  As shown in \autoref{fig:vdr_stats}(b), we compute the token-length distribution of the natural-language questions. The average question length is 25.54 tokens, and the lengths are concentrated: the 95th percentile is 54 tokens. This indicates that most questions are compact in phrasing, maintaining concise language while preserving necessary constraints. As a result, evaluation can focus more on the core process of starting from video cues and combining web evidence to perform multi-hop reasoning, rather than on extra comprehension burden induced by overly long inputs.


\paragraph{Video Duration.} As shown in \autoref{fig:vdr_stats}(c), the video durations exhibit a long-tailed distribution, covering both short clips and substantially longer videos. Based on our duration-based grouping used in later stratified experiments, the benchmark contains short, medium, and long videos in comparable proportions: 32.4\% Short, 30.2\% Medium, and 37.4\% Long. This duration structure covers both rapid cue capture in short-video scenarios and cross-segment association in long-video scenarios, enabling a more comprehensive test of models' cross-frame understanding and evidence localization across different time scales.

\section{Experiments}

\subsection{Setting}

\textbf{Baselines.} We compare two common paradigms: Workflow and Agentic. 
Workflow adopts a two-stage design: a multimodal model first extracts question-relevant 
cross-frame visual cues from the video and produces a structured intermediate text, 
which is then provided together with the question as input for subsequent reasoning. 
Without further access to the original video, the system uses the search tool to retrieve 
candidate evidence from the open web, and performs explicit reflection and evidence 
filtering to generate the next-round query, finally aggregating multi-round search results 
to produce an answer. In contrast, Agentic adopts a stronger end-to-end setting: it feeds 
the raw video and the question directly into a single multimodal agent, which performs 
video understanding, query generation, web retrieval, and evidence integration within 
the same execution loop. The agent uses the search tool as its external interface to the 
open web, while autonomously deciding when to initiate search based on video cues, 
how to refine its queries through intermediate reasoning, and how to organize multi-round 
evidence into the final conclusion.

\textbf{MLLMs.} We select mainstream multimodal models spanning both closed-source and open-source families as the core research agents. Closed-source models include Gemini-3.1-pro-preview~\cite{google2026gemini31propreview} and GPT-5.4~\cite{openai2026gpt54}; open-source models include Qwen3.5-35B-A3B~\cite{qwen2026qwen35}, InternVL3.5-14B~\cite{wang2025internvl35}, Gemma-4-31B-it~\cite{googledeepmind2026gemma4}, and GLM-4.6V-Flash~\cite{zai2025glm46vflash}. All models are tested under both Workflow and Agentic paradigms to analyze their capability boundaries under different system organizations.

\textbf{LLM as Judge.} \method{} is an open-domain factual question answering task. To avoid false mismatches, we adopt an LLM-as-judge protocol \cite{zheng2023judging} using DeepSeek-V3-0324 \cite{liu2024deepseek} to assess semantic equivalence between the model prediction and the reference answer. The judge outputs a binary correctness label, which we use to compute overall Accuracy.

\subsection{Main Results}   



\begin{table*}[t]
\centering
\caption{Performance comparison across difficulty levels.}
\label{tab:model-performance}

\begingroup
\footnotesize
\setlength{\tabcolsep}{3.2pt}
\renewcommand{\arraystretch}{0.96}
\setlength{\aboverulesep}{0.35ex}
\setlength{\belowrulesep}{0.35ex}
\setlength{\cmidrulesep}{0.15ex}

\begin{tabular*}{\textwidth}{@{\extracolsep{\fill}}lcccccccc@{}}
\toprule
\multirow{2}{*}{\textbf{Model}} 
& \multicolumn{4}{c}{\textbf{Workflow}} 
& \multicolumn{4}{c}{\textbf{Agentic}} \\
\cmidrule(lr){2-5} 
\cmidrule(lr){6-9}
& \textbf{Low} 
& \textbf{Mid} 
& \textbf{High} 
& \textbf{Ave.} 
& \textbf{Low} 
& \textbf{Mid} 
& \textbf{High} 
& \textbf{Ave.} \\
\midrule
\textbf{\#Samples} 
& 112 & 171 & 217 & 500 
& 112 & 171 & 217 & 500 \\
\midrule
Qwen3.5-35B-A3B        
& 61.61 & 37.43 & 19.35 & 35.00 
& 57.14 & 40.94 & 17.97 & 34.60 \\
InternVL3.5-14B        
& 29.46 & 16.96 &  8.29 & 16.00 
& 30.36 & 16.96 &  6.45 & 15.40 \\
Gemma-4-31B-it         
& 52.68 & 38.60 & 15.21 & 31.60 
& 57.14 & 43.27 & 13.82 & 33.60 \\
Gemini-3.1-pro-preview 
& 78.57 & 63.74 & 28.57 & 51.80 
& 84.82 & 62.57 & 30.41 & 53.60 \\
GPT-5.4                
& 66.96 & 52.63 & 32.26 & 47.00 
& 71.43 & 53.80 & 29.95 & 47.40 \\
GLM-4.6V-Flash         
& 33.93 & 19.30 &  8.29 & 17.80 
& 43.75 & 23.98 &  8.76 & 21.80 \\
\midrule
Human                  
& 81.71 & 49.64 & 11.39 & 40.22 
& 81.71 & 49.64 & 11.39 & 40.22 \\
\bottomrule
\end{tabular*}

\endgroup
\end{table*}
\begin{table*}[t]
\centering
\caption{Performance comparison across video length groups.}
\label{tab:model-performance-length}

\begingroup
\footnotesize
\setlength{\tabcolsep}{3.2pt}
\renewcommand{\arraystretch}{0.96}
\setlength{\aboverulesep}{0.35ex}
\setlength{\belowrulesep}{0.35ex}
\setlength{\cmidrulesep}{0.15ex}

\begin{tabular*}{\textwidth}{@{\extracolsep{\fill}}lcccccccc@{}}
\toprule
\multirow{2}{*}{\textbf{Model}} 
& \multicolumn{4}{c}{\textbf{Workflow}} 
& \multicolumn{4}{c}{\textbf{Agentic}} \\
\cmidrule(lr){2-5} 
\cmidrule(lr){6-9}
& \textbf{Short} 
& \textbf{Medium} 
& \textbf{Long} 
& \textbf{Ave.} 
& \textbf{Short} 
& \textbf{Medium} 
& \textbf{Long} 
& \textbf{Ave.} \\
\midrule
\textbf{\#Samples} 
& 162 & 151 & 187 & 500 
& 162 & 151 & 187 & 500 \\
\midrule
Qwen3.5-35B-A3B        
& 30.86 & 35.76 & 37.97 & 35.00 
& 31.48 & 33.77 & 37.97 & 34.60 \\
InternVL3.5-14B        
& 17.28 & 11.92 & 18.18 & 16.00 
& 12.35 & 15.89 & 17.65 & 15.40 \\
Gemma-4-31B-it         
& 23.46 & 25.17 & 43.85 & 31.60 
& 30.25 & 29.80 & 39.57 & 33.60 \\
Gemini-3.1-pro-preview 
& 51.85 & 49.01 & 54.01 & 51.80 
& 54.32 & 47.02 & 58.29 & 53.60 \\
GPT-5.4                
& 42.59 & 43.71 & 53.48 & 47.00 
& 43.83 & 40.40 & 56.15 & 47.40 \\
GLM-4.6V-Flash         
& 15.43 & 18.54 & 19.25 & 17.80 
& 20.37 & 23.84 & 21.39 & 21.80 \\
\midrule
Human                  
& 41.65 & 37.91 & 40.85 & 40.22 
& 41.65 & 37.91 & 40.85 & 40.22 \\
\bottomrule
\end{tabular*}

\endgroup
\end{table*}
\begin{table*}[t]
\centering
\caption{Performance comparison across different domains.}
\label{tab:model-performance-domains}

\begingroup
\footnotesize
\setlength{\tabcolsep}{2.6pt}
\renewcommand{\arraystretch}{0.94}
\setlength{\aboverulesep}{0.35ex}
\setlength{\belowrulesep}{0.35ex}
\setlength{\cmidrulesep}{0.12ex}

\begin{tabular*}{\textwidth}{@{\extracolsep{\fill}}llccccccc@{}}
\toprule
\multirow{2}{*}{\textbf{Model}} 
& \multirow{2}{*}{\textbf{Setting}} 
& \multicolumn{7}{c}{\textbf{Domain (\%)}} \\
\cmidrule(lr){3-9}
& 
& \textbf{History} 
& \textbf{Geography} 
& \textbf{Culture} 
& \textbf{Economics} 
& \textbf{Technology} 
& \textbf{Daily Life} 
& \textbf{Ave.} \\
\midrule
\multicolumn{2}{l}{\textbf{\#Samples}} 
& 59 & 38 & 65 & 57 & 113 & 168 & 500 \\
\midrule
\multirow{2}{*}{Qwen3.5-35B-A3B}
& Workflow 
& 32.20 & 28.95 & 32.31 & 22.81 & 30.09 & 45.83 & 35.00 \\
& Agentic  
& 28.81 & 28.95 & 27.69 & 35.09 & 34.51 & 40.48 & 34.60 \\
\cmidrule(lr){2-9}
\multirow{2}{*}{InternVL3.5-14B}
& Workflow 
& 18.64 &  5.26 & 15.38 & 12.28 & 15.93 & 19.05 & 16.00 \\
& Agentic  
& 18.64 &  2.63 & 20.00 & 24.56 &  6.19 & 18.45 & 15.40 \\
\cmidrule(lr){2-9}
\multirow{2}{*}{Gemma-4-31B-it}
& Workflow 
& 23.73 & 28.95 & 26.15 & 19.30 & 30.09 & 42.26 & 31.60 \\
& Agentic  
& 27.12 & 18.42 & 32.31 & 24.56 & 37.17 & 40.48 & 33.60 \\
\cmidrule(lr){2-9}
\multirow{2}{*}{Gemini-3.1-pro-preview}
& Workflow 
& 52.54 & 47.37 & 49.23 & 42.11 & 50.44 & 57.74 & 51.80 \\
& Agentic  
& 59.32 & 47.37 & 49.23 & 35.09 & 54.87 & 60.12 & 53.60 \\
\cmidrule(lr){2-9}
\multirow{2}{*}{GPT-5.4}
& Workflow 
& 52.54 & 44.74 & 52.31 & 33.33 & 39.82 & 52.98 & 47.00 \\
& Agentic  
& 47.46 & 55.26 & 46.15 & 29.82 & 46.02 & 52.98 & 47.40 \\
\cmidrule(lr){2-9}
\multirow{2}{*}{GLM-4.6V-Flash}
& Workflow 
& 16.95 & 10.53 & 20.00 & 12.28 & 21.24 & 18.45 & 17.80 \\
& Agentic  
& 20.34 & 10.53 & 27.69 & 21.05 & 18.58 & 25.00 & 21.80 \\
\midrule
Human 
& 
& 39.62 & 37.97 & 47.81 & 39.53 & 37.94 & 39.77 & 40.22 \\
\bottomrule
\end{tabular*}

\endgroup
\end{table*}

From the overall averages, our \method{} setting yields a clear capability stratification among existing models. Gemini-3.1-pro-preview leads under both paradigms, achieving 51.80\% under Workflow and 53.60\% under Agentic. GPT-5.4 demonstrates the second-best overall performance, with 47.00\% under Workflow and 47.40\% under Agentic. Qwen3.5-35B-A3B and Gemma-4-31B-it form a middle tier, with Qwen3.5-35B-A3B achieving 35.00\% / 34.60\% and Gemma-4-31B-it achieving 31.60\% / 33.60\% under Workflow / Agentic, respectively. InternVL3.5-14B and GLM-4.6V-Flash are overall weaker, with InternVL3.5-14B reaching 16.00\% / 15.40\% and GLM-4.6V-Flash reaching 17.80\% / 21.80\%. Compared with human performance (40.22\%), only Gemini-3.1-pro-preview and GPT-5.4 consistently surpass the human average under this evaluation setting.

\textbf{Difficulty stratification.} This reveals the essential differences between the two paradigms. As shown in \autoref{tab:model-performance}, all evaluated models exhibit a highly consistent decline in performance as the difficulty level increases from Low to Mid and then to High. Human performance decreases from 81.71\% to 49.64\% and then to 11.39\%, and models also generally degrade as difficulty increases, suggesting that our difficulty definition based on human success rate indeed corresponds to longer and more fragile evidence chains. However, the effect of Agentic varies substantially across models and difficulty levels. For example, Gemini-3.1-pro-preview improves on Low and High difficulty levels, from 78.57\%$\rightarrow$84.82\% and 28.57\%$\rightarrow$30.41\%, respectively, while slightly dropping on Mid from 63.74\% to 62.57\%. GPT-5.4 improves on Low/Mid (66.96\%$\rightarrow$71.43\%, 52.63\%$\rightarrow$53.80\%), but drops on High (32.26\%$\rightarrow$29.95\%). Qwen3.5-35B-A3B and InternVL3.5-14B also show declines on High when switching to Agentic, from 19.35\%$\rightarrow$17.97\% and 8.29\%$\rightarrow$6.45\%, respectively. In contrast, GLM-4.6V-Flash improves slightly on High from 8.29\% to 8.76\%, although its absolute performance remains low. These results indicate that directly placing the video inside the agent loop does not automatically improve difficult cases. This phenomenon can be partly explained by our implementation constraints: in Agentic, the model cannot re-watch the video during research, so the subsequent process must rely solely on cues obtained from the initial viewing to guide multi-round search and reasoning, lacking opportunities for online correction via revisiting the video. Once early visual anchors drift, web noise can amplify the deviation in later searches, making High-difficulty questions more prone to goal drift; in contrast, Workflow externalizes video cues into an intermediate textual representation, effectively providing a repeatedly accessible external memory that reduces drift risk in long-horizon decision-making.

\textbf{Video-duration stratification.} This further reinforces the above observation: as videos get longer, cues become more dispersed and the state space grows, so Agentic places higher demands on a model's ability to retain and continuously leverage initial video cues, leading to strong polarization. As shown in \autoref{tab:model-performance-length}, Gemini-3.1-pro-preview gains on Short and Long videos, improving from 51.85\%$\rightarrow$54.32\% and 54.01\%$\rightarrow$58.29\%, respectively, while dropping on Medium videos from 49.01\% to 47.02\%. GPT-5.4 shows a similar mixed pattern, improving on Short and Long videos (42.59\%$\rightarrow$43.83\%, 53.48\%$\rightarrow$56.15\%) but dropping on Medium videos (43.71\%$\rightarrow$40.40\%). For weaker or less stable models, the effect is also heterogeneous: Qwen3.5-35B-A3B slightly improves on Short videos, drops on Medium videos, and remains unchanged on Long videos; InternVL3.5-14B drops on Short and Long videos but improves on Medium videos; Gemma-4-31B-it improves on Short and Medium videos but drops on Long videos from 43.85\% to 39.57\%. GLM-4.6V-Flash improves across all duration groups, but its absolute accuracy remains much lower than that of the leading models. These results suggest that Agentic does not uniformly benefit from longer or richer video input: when a model struggles to preserve and reuse key video anchors across multiple rounds of search and reasoning, direct video access may fail to improve search decisions and can even destabilize long reasoning chains. In other words, long videos amplify the trade-off between the two paradigms: Workflow may lose some visual details but provides more controllable textual anchors, whereas Agentic has direct access to raw video information but depends more on long-term state consistency.

\textbf{Domain stratification.} This highlights the respective advantages of the two paradigms across domains. As shown in \autoref{tab:model-performance-domains}, Gemini-3.1-pro-preview benefits from Agentic in History, Technology, and Daily Life, improving from 52.54\%$\rightarrow$59.32\%, 50.44\%$\rightarrow$54.87\%, and 57.74\%$\rightarrow$60.12\%, respectively. It remains unchanged in Geography and Culture, while dropping in Economics from 42.11\% to 35.09\%. GPT-5.4 shows a different pattern: it improves in Geography and Technology (44.74\%$\rightarrow$55.26\%, 39.82\%$\rightarrow$46.02\%), remains unchanged in Daily Life, but drops in History, Culture, and Economics. Other models also exhibit domain-specific trade-offs. For instance, Gemma-4-31B-it improves in History, Culture, Economics, and Technology, but drops in Geography and Daily Life; Qwen3.5-35B-A3B improves in Economics and Technology, but drops in History, Culture, and Daily Life. This domain-level variation further shows that the benefit of Agentic is not universal, but depends jointly on the model's video-memory stability and the domain-specific ambiguity of downstream search. Domains that require fine-grained visual cues to be translated into precise search constraints may benefit from direct video access when the underlying model is strong enough, whereas domains with ambiguous or weak visual anchors remain vulnerable to search drift.

Across all three stratifications, we conclude that under current mainstream model capabilities, Agentic is not necessarily superior to Workflow: its effectiveness depends on whether the model can continually rely on initial video cues throughout multi-round search and reasoning; meanwhile, Workflow provides stable anchors for downstream search and reasoning through an explicit intermediate text, thereby more effectively reducing the risk that the search process drifts away from the correct target.

\begin{table*}[t]
\centering
\caption{Error type distribution under both the \textbf{Workflow} and the \textbf{Agentic} settings.}
\label{tab:error-type-distribution}

\begingroup
\scriptsize
\setlength{\tabcolsep}{2.3pt}
\renewcommand{\arraystretch}{0.94}
\setlength{\aboverulesep}{0.35ex}
\setlength{\belowrulesep}{0.35ex}
\setlength{\cmidrulesep}{0.12ex}

\begin{tabular*}{\textwidth}{@{\extracolsep{\fill}}llccccccccc@{}}
\toprule
\multirow{2}{*}{\textbf{Model}} 
& \multirow{2}{*}{\textbf{Setting}} 
& \multicolumn{9}{c}{\textbf{Error Type Count}} \\
\cmidrule(lr){3-11}
& 
& \textbf{Categorical} 
& \textbf{Incomplete} 
& \textbf{Not Found} 
& \textbf{Numerical} 
& \textbf{Context} 
& \textbf{Semantic} 
& \textbf{Reasoning} 
& \textbf{Others} 
& \textbf{Total} \\
\midrule
\multirow{2}{*}{Qwen3.5-35B-A3B}
& Workflow 
& 129 & 22 & 21 & 110 & 20 & 10 & 13 &  0 & 325 \\
& Agentic  
& 114 & 25 & 24 & 114 & 35 & 10 &  5 &  0 & 327 \\
\cmidrule(lr){2-11}
\multirow{2}{*}{InternVL3.5-14B}
& Workflow 
& 155 & 31 & 56 & 141 & 15 & 14 &  6 &  2 & 420 \\
& Agentic  
& 150 & 26 & 32 & 156 & 26 & 17 &  7 &  9 & 423 \\
\cmidrule(lr){2-11}
\multirow{2}{*}{Gemma-4-31B-it}
& Workflow 
& 115 &  1 & 95 &  91 & 16 & 10 &  2 & 12 & 342 \\
& Agentic  
& 118 &  6 & 76 &  95 &  4 & 15 & 10 &  8 & 332 \\
\cmidrule(lr){2-11}
\multirow{2}{*}{Gemini-3.1-pro-preview}
& Workflow 
&  79 & 27 & 15 &  84 & 13 & 17 &  4 &  2 & 241 \\
& Agentic  
&  86 & 16 &  4 &  95 & 15 & 13 &  3 &  0 & 232 \\
\cmidrule(lr){2-11}
\multirow{2}{*}{GPT-5.4}
& Workflow 
& 103 & 17 & 15 & 109 &  9 &  8 &  2 &  2 & 265 \\
& Agentic  
&  99 & 17 & 11 & 123 &  4 &  8 &  1 &  0 & 263 \\
\cmidrule(lr){2-11}
\multirow{2}{*}{GLM-4.6V-Flash}
& Workflow 
&  15 & 52 & 53 & 106 & 42 &  5 & 101 & 37 & 411 \\
& Agentic  
&  25 & 58 & 38 & 109 & 41 &  7 &  96 & 17 & 391 \\
\bottomrule
\end{tabular*}

\endgroup
\end{table*}

\subsection{Error Analysis}

To obtain a more fine-grained understanding of MLLMs' behaviors under different paradigms, we conduct a detailed analysis of their error trajectories across all experiments. We further break down the errors into eight categories, as shown in \autoref{tab:error-type-distribution}.

\textbf{Perception Grounding Limitation.} \autoref{tab:error-type-distribution} indicates that Categorical Error remains a major failure mode for many models, especially Qwen3.5-35B-A3B, InternVL3.5-14B, and Gemma-4-31B-it, while Numerical Error is comparably prominent for stronger models such as Gemini-3.1-pro-preview and GPT-5.4. Under Workflow, categorical errors range from 79 to 155 among these representative models, and under Agentic, they remain high, ranging from 86 to 150. This pattern is closely tied to our setting: once the research phase cannot revisit the video, downstream retrieval and reasoning depend heavily on the visual anchors extracted from the first pass. When early perception deviates, there is no mechanism to re-localize key frames and correct the anchor, making error propagation along the evidence chain more likely. At the same time, the error distribution also reveals model-specific weaknesses. For example, GLM-4.6V-Flash has an unusually large number of Reasoning Errors under both Workflow and Agentic settings (101 and 96), indicating that its failures are not limited to perception grounding, but also involve evidence integration and reasoning instability. This further suggests that different models fail at different stages of the overall video-grounded open-domain QA pipeline.

\textbf{Numerical Error Bottleneck.} \autoref{tab:error-type-distribution} shows that Gemini-3.1-pro-preview has the lowest overall error count among all models, with 241 total errors under Workflow and 232 under Agentic, yet this advantage does not eliminate Numerical Error. Under Workflow, Gemini-3.1-pro-preview still has 84 numerical errors, while Qwen3.5-35B-A3B, InternVL3.5-14B, Gemma-4-31B-it, GPT-5.4, and GLM-4.6V-Flash have 110, 141, 91, 109, and 106, respectively. Under Agentic, Gemini-3.1-pro-preview records 95 numerical errors, compared with 114, 156, 95, 123, and 109 for the same five models. Numerical errors therefore remain substantial across all evaluated models, even for the strongest systems. This suggests that numerical reliability constitutes a distinct and persistent weakness for current MLLMs: while stronger models can reduce not-found errors and improve evidence grounding, they still frequently fail when the task requires precise counting, comparison, or numerical verification.

\section{Conclusion}

We introduce \method{}, a video deep research benchmark for evaluating multimodal deep research on the open web. It requires models to extract multi-frame visual anchors, turn them into searchable queries, retrieve open-web evidence, and perform multi-hop reasoning to produce verifiable factual answers. We benchmark mainstream multimodal models under Workflow and Agentic paradigms, with stratified analyses by difficulty, video duration, and semantic domain to characterize capability boundaries on video deep search.



\bibliography{anthology}

\newpage
\appendix
\newcommand{\VideoDRCodeURL}{https://anonymous.4open.science/r/VideoDR-Benchmark-B345/}
\newcommand{\VideoDRDataURL}{https://anonymous-hf.up.railway.app/a/kua2plysxsqt/}

\section{Retrieval-and-Reasoning Analysis}

As shown in \autoref{tab:tool-use}, the number of search calls and reasoning steps is not monotonically correlated with accuracy. What determines performance is not using search more, but whether a model can turn a limited number of search calls into a high-quality evidence chain. Concretely, Gemini-3.1-pro-preview demonstrates stronger retrieval-and-reasoning effectiveness: under the Agentic setting, it averages 4.55/3.16 reasoning/search calls with 345.73s runtime and achieves the best overall score of 53.60\%, indicating that additional retrieval and reflection are converted into more reliable evidence integration for this model. In contrast, GLM-4.6V-Flash consumes the longest runtime under both paradigms, with 511.98s in Workflow and 502.96s in Agentic, but only achieves 17.80\% / 21.80\% accuracy, suggesting that longer execution time does not necessarily lead to better evidence acquisition or answer correctness. Qwen3.5-35B-A3B also shows that reducing reasoning and search calls does not automatically improve or harm performance in a simple way: its Agentic setting uses fewer reasoning/search calls than Workflow, 1.59/3.02 vs. 2.97/3.80, and runs faster, 374.47s vs. 485.57s, but accuracy still slightly drops from 35.00\% to 34.60\%. Gemma-4-31B-it provides another contrasting case: Agentic improves accuracy from 31.60\% to 33.60\%, while increasing reasoning calls from 3.69 to 3.84, reducing search calls from 2.00 to 1.70, and increasing runtime from 232.28s to 285.67s. These observations suggest that interaction quality, rather than raw interaction frequency or runtime, is the key factor determining whether additional search and reflection can improve performance.

\begin{table*}[t]
\centering
\caption{Retrieval-and-reasoning statistics under the \textbf{Workflow} and \textbf{Agentic} settings.}
\label{tab:tool-use}

\begingroup
\footnotesize
\setlength{\tabcolsep}{3.2pt}
\renewcommand{\arraystretch}{0.96}
\setlength{\aboverulesep}{0.35ex}
\setlength{\belowrulesep}{0.35ex}
\setlength{\cmidrulesep}{0.15ex}

\begin{tabular*}{0.8\textwidth}{@{\extracolsep{\fill}}lcccccc@{}}
\toprule
\multirow{2}{*}{\textbf{Model}} 
& \multicolumn{3}{c}{\textbf{Workflow}} 
& \multicolumn{3}{c}{\textbf{Agentic}} \\
\cmidrule(lr){2-4}
\cmidrule(lr){5-7}
& \textbf{reasoning} 
& \textbf{\textbf{search}} 
& \textbf{time (s)} 
& \textbf{reasoning} 
& \textbf{\textbf{search}} 
& \textbf{time (s)} \\
\midrule
Qwen3.5-35B-A3B     
& 2.97 & 3.80 & 485.57 
& 1.59 & 3.02 & 374.47 \\
InternVL3.5-14B     
& 2.22 & 2.84 & 154.03 
& 1.86 & 2.49 & 126.37 \\
Gemma-4-31B-it      
& 3.69 & 2.00 & 232.28 
& 3.84 & 1.70 & 285.67 \\
Gemini-3.1-pro-preview 
& 3.58 & 2.30 & 188.20 
& 4.55 & 3.16 & 345.73 \\
GPT-5.4             
& 3.88 & 2.86 & 178.84 
& 3.84 & 2.78 & 166.32 \\
GLM-4.6V-Flash      
& 2.05 & 2.40 & 511.98 
& 1.92 & 2.39 & 502.96 \\
\bottomrule
\end{tabular*}

\endgroup
\end{table*}

\section{Token Usage Analysis}

As shown in \autoref{tab:token-usage}, Agentic generally introduces higher token consumption than Workflow, especially in prompt tokens, because the model directly processes the raw video-conditioned context and autonomously maintains the retrieval-and-reasoning trajectory within a single execution loop. This trend is particularly evident for Qwen3.5-35B-A3B, whose average total tokens increase from 168,483 to 301,187, and for Gemini-3.1-pro-preview, whose average total tokens increase from 80,271 to 199,008. GLM-4.6V-Flash also shows a substantial increase from 56,839 to 96,329 total tokens. In contrast, GPT-5.4 is the only model whose Agentic setting slightly reduces average total token usage, from 109,236 to 101,936, suggesting more compact interaction behavior under the end-to-end agent loop. Across models, completion tokens remain relatively stable or even decrease in several Agentic settings, while prompt tokens dominate the overall increase, indicating that the main efficiency burden of Agentic comes from maintaining larger contextual states rather than generating longer responses. The high P95 values, especially for Qwen3.5-35B-A3B under Agentic, further suggest that Agentic can occasionally trigger very long search-and-reasoning trajectories, making token cost less predictable than Workflow.

\begin{table*}[t]
\centering
\caption{Token usage statistics under the \textbf{Workflow} and \textbf{Agentic} settings.}
\label{tab:token-usage}

\begingroup
\scriptsize
\setlength{\tabcolsep}{2.4pt}
\renewcommand{\arraystretch}{0.94}
\setlength{\aboverulesep}{0.35ex}
\setlength{\belowrulesep}{0.35ex}
\setlength{\cmidrulesep}{0.12ex}

\begin{tabular*}{\textwidth}{@{\extracolsep{\fill}}llrrrrr@{}}
\toprule
\textbf{Model} 
& \textbf{Setting} 
& \begin{tabular}{@{}c@{}}\textbf{Avg Prompt}\\\textbf{Tokens}\end{tabular}
& \begin{tabular}{@{}c@{}}\textbf{Avg Completion}\\\textbf{Tokens}\end{tabular}
& \begin{tabular}{@{}c@{}}\textbf{Avg Total}\\\textbf{Tokens}\end{tabular}
& \begin{tabular}{@{}c@{}}\textbf{Median Total}\\\textbf{Tokens}\end{tabular}
& \begin{tabular}{@{}c@{}}\textbf{P95 Total}\\\textbf{Tokens}\end{tabular} \\
\midrule
\multirow{2}{*}{Qwen3.5-35B-A3B}
& Workflow & 159621 & 8861 & 168483 & 163890 & 337718 \\
& Agentic  & 295004 & 6183 & 301187 & 200927 & 755463 \\
\cmidrule(lr){2-7}
\multirow{2}{*}{InternVL3.5-14B}
& Workflow &  53209 & 7481 &  60690 &  53202 & 123752 \\
& Agentic  &  59108 & 6379 &  65487 &  58544 & 131373 \\
\cmidrule(lr){2-7}
\multirow{2}{*}{Gemma-4-31B-it}
& Workflow &  54814 & 5458 &  60272 &  37538 & 174100 \\
& Agentic  &  70917 & 5676 &  76592 &  60790 & 195298 \\
\cmidrule(lr){2-7}
\multirow{2}{*}{Gemini-3.1-pro-preview}
& Workflow &  71949 & 8322 &  80271 &  66936 & 153754 \\
& Agentic  & 189733 & 9275 & 199008 & 153038 & 364158 \\
\cmidrule(lr){2-7}
\multirow{2}{*}{GPT-5.4}
& Workflow & 103461 & 5775 & 109236 &  99682 & 187596 \\
& Agentic  &  96933 & 5003 & 101936 &  94815 & 169805 \\
\cmidrule(lr){2-7}
\multirow{2}{*}{GLM-4.6V-Flash}
& Workflow &  50289 & 6550 &  56839 &  47409 & 132597 \\
& Agentic  &  90545 & 5784 &  96329 &  83331 & 203463 \\
\bottomrule
\end{tabular*}

\endgroup
\end{table*}

\section{Efficiency Analysis}

As shown in \autoref{tab:efficiency-statistics}, higher accuracy does not necessarily imply higher efficiency, and the relative efficiency of Workflow and Agentic varies substantially across models. Gemini-3.1-pro-preview achieves the best accuracy under Agentic, improving from 51.80\% to 53.60\%, but this gain comes with much higher average runtime, increasing from 188.20s to 345.73s, and lower token efficiency, dropping from 0.65 to 0.27 accuracy per 1K tokens. A similar efficiency trade-off appears for Gemma-4-31B-it, where Agentic improves accuracy from 31.60\% to 33.60\% but increases both runtime and tokens per correct sample. By contrast, GPT-5.4 shows the most favorable Agentic efficiency profile and remains relatively stable across the two paradigms: Agentic slightly improves accuracy from 47.00\% to 47.40\%, while reducing average runtime from 178.84s to 166.32s and improving both accuracy per 1K tokens and accuracy per minute. For weaker models, Agentic often improves speed but not necessarily cost-effectiveness; for example, Qwen3.5-35B-A3B runs faster under Agentic but requires far more tokens per correct sample, increasing from 481,379 to 870,483. Overall, these results reinforce that Agentic is not uniformly more efficient than Workflow: its practical value depends on whether the additional context and autonomous reasoning process can be converted into accuracy gains without excessive token or latency overhead.

\begin{table*}[t]
\centering
\caption{Efficiency statistics under the \textbf{Workflow} and \textbf{Agentic} settings.}
\label{tab:efficiency-statistics}

\begingroup
\scriptsize
\setlength{\tabcolsep}{2.1pt}
\renewcommand{\arraystretch}{0.94}
\setlength{\aboverulesep}{0.35ex}
\setlength{\belowrulesep}{0.35ex}
\setlength{\cmidrulesep}{0.12ex}

\begin{tabular*}{\textwidth}{@{\extracolsep{\fill}}llrrrrrrr@{}}
\toprule
\textbf{Model} 
& \textbf{Setting} 
& \begin{tabular}{@{}c@{}}\textbf{Judge}\\\textbf{Accuracy}\end{tabular}
& \begin{tabular}{@{}c@{}}\textbf{Avg Time}\\\textbf{(s)}\end{tabular}
& \begin{tabular}{@{}c@{}}\textbf{Median Time}\\\textbf{(s)}\end{tabular}
& \begin{tabular}{@{}c@{}}\textbf{P95 Time}\\\textbf{(s)}\end{tabular}
& \begin{tabular}{@{}c@{}}\textbf{Acc /}\\\textbf{1K Tokens}\end{tabular}
& \begin{tabular}{@{}c@{}}\textbf{Acc /}\\\textbf{Min}\end{tabular}
& \begin{tabular}{@{}c@{}}\textbf{Tokens /}\\\textbf{Correct Sample}\end{tabular} \\
\midrule
\multirow{2}{*}{Qwen3.5-35B-A3B}
& Workflow & 35.00 & 485.57 & 382.97 & 1300.26 & 0.21 &  4.32 & 481378.61 \\
& Agentic  & 34.60 & 374.47 & 284.48 &  991.93 & 0.11 &  5.54 & 870482.53 \\
\cmidrule(lr){2-9}
\multirow{2}{*}{InternVL3.5-14B}
& Workflow & 16.00 & 154.03 & 124.35 &  349.50 & 0.26 &  6.23 & 379312.29 \\
& Agentic  & 15.40 & 126.37 & 111.28 &  303.93 & 0.24 &  7.31 & 425239.98 \\
\cmidrule(lr){2-9}
\multirow{2}{*}{Gemma-4-31B-it}
& Workflow & 31.60 & 232.28 & 172.94 &  567.54 & 0.52 &  8.16 & 190733.88 \\
& Agentic  & 33.60 & 285.67 & 197.42 &  854.56 & 0.44 &  7.06 & 227953.85 \\
\cmidrule(lr){2-9}
\multirow{2}{*}{Gemini-3.1-pro-preview}
& Workflow & 51.80 & 188.20 & 149.44 &  365.58 & 0.65 & 16.51 & 154963.72 \\
& Agentic  & 53.60 & 345.73 & 300.30 &  783.26 & 0.27 &  9.30 & 371283.09 \\
\cmidrule(lr){2-9}
\multirow{2}{*}{GPT-5.4}
& Workflow & 47.00 & 178.84 & 139.70 &  356.98 & 0.43 & 15.77 & 231414.84 \\
& Agentic  & 47.40 & 166.32 & 131.35 &  339.31 & 0.47 & 17.10 & 214491.07 \\
\cmidrule(lr){2-9}
\multirow{2}{*}{GLM-4.6V-Flash}
& Workflow & 17.80 & 511.98 & 402.86 & 1224.30 & 0.31 &  2.09 & 319319.05 \\
& Agentic  & 21.80 & 502.96 & 362.78 & 1283.71 & 0.23 &  2.60 & 441875.81 \\
\bottomrule
\end{tabular*}

\endgroup
\end{table*}

\section{Annotation Effort}

Constructing VideoDR requires substantial manual effort because each sample must satisfy
both video dependency and web dependency. Annotators need to inspect videos repeatedly,
identify multi-frame visual anchors, design questions that require multi-hop open-web
verification, archive supporting evidence, and further filter out samples answerable by
video-only or web-only shortcuts. Although this process is time-consuming and limits rapid
scaling, it is necessary to ensure that each retained question has a unique, verifiable
answer and genuinely evaluates video-grounded deep research rather than simple video
understanding or text-based web search.

\section{Data}
\label{sec:data-code-repro}



We also prepare an open-source dataset release at:
\begin{center}
\url{https://huggingface.co/datasets/strike20023/VideoDR}
\end{center}

The dataset release contains the benchmark annotations and the structured fields needed to reconstruct benchmark samples, including the question-answer pairs, video references, evidence references, and sample-level metadata used for evaluation.

To facilitate reproducibility and long-term auditing, we preserve local copies of all videos and evidence webpages involved in annotation and verification. Each benchmark item is associated with source references and archived evidence records. In the public release, we focus on benchmark annotations, metadata, source URLs, evidence references, and evaluation scripts. Third-party raw videos or webpages are redistributed only when their licenses or terms of use explicitly permit such redistribution.

\section{Previous Version Experimental Results}
\label{sec:appendix-v1}

This section reports the experimental tables from the previous version (v1) of this manuscript.

\subsection{Overall Performance by Difficulty}

\autoref{tab:v1-model-performance} reports the v1 accuracy results under the Workflow and Agentic settings across Low, Mid, and High difficulty groups. In the v1 setting, difficulty was also defined according to human success rate, and the table was used to analyze how model performance changes as the evidence chain becomes longer and more fragile.

\begin{table*}[t]
\centering
\small
\setlength{\tabcolsep}{5pt}
\renewcommand{\arraystretch}{1.3}
\caption{Performance comparison across difficulty levels under the \textbf{Workflow} and the \textbf{Agentic} settings.}
\begin{tabular}{lcccccccc}
\toprule
\multirow{2}{*}{\textbf{Model}} & \multicolumn{4}{c}{\textbf{Workflow}} & \multicolumn{4}{c}{\textbf{Agentic}} \\
\cmidrule(lr){2-5} \cmidrule(lr){6-9}
 & \textbf{Low} & \textbf{Mid} & \textbf{High} & \textbf{Ave.} & \textbf{Low} & \textbf{Mid} & \textbf{High} & \textbf{Ave.} \\
\midrule
\textbf{\#Samples} & 32 & 36 & 32 & 100 & 32 & 36 & 32 & 100 \\
\midrule

Qwen3-Omni-30B-A3B & 59.38 & 30.56 & 21.88 & 37.00 & 65.62 & 27.78 & 18.75 & 37.00 \\
InternVL3.5-14B    & 37.50 & 22.22 & 21.88 & 27.00 & 46.88 & 22.22 & 21.88 & 30.00 \\
MiniCPM-V 4.5      & 50.00 & 13.89 & 12.50 & 25.00 & 18.75 & 19.44 &  9.38 & 16.00 \\
Gemini-3-pro-preview & 90.62 & 61.11 & 56.25 & 69.00 & 93.75 & 69.44 & 65.62 & 76.00 \\
GPT-4o             & 50.00 & 30.56 & 46.88 & 42.00 & 62.50 & 38.89 & 28.12 & 43.00 \\
GPT-5.2            & 84.38 & 61.11 & 62.50 & 69.00 & 84.38 & 58.33 & 65.62 & 69.00 \\
\midrule

Human              & 90.00 & 50.56 & 10.63 & 50.40 & 90.00 & 50.56 & 10.63 & 50.40 \\
\bottomrule
\end{tabular}
\label{tab:v1-model-performance}
\end{table*}

\subsection{Performance by Video Duration}

\autoref{tab:v1-model-performance-duration} reports the v1 results grouped by video duration. This table studyed whether longer videos make it harder for models to preserve and reuse the initial visual anchors during subsequent web search and reasoning.

\begin{table*}[t]
\centering
\small
\setlength{\tabcolsep}{5pt}
\renewcommand{\arraystretch}{1.2}
\caption{Performance comparison across different video durations under the \textbf{Workflow} and the \textbf{Agentic} settings.}
\begin{tabular}{lcccccccc}
\toprule
\multirow{2}{*}{\textbf{Model}} & \multicolumn{4}{c}{\textbf{Workflow}} & \multicolumn{4}{c}{\textbf{Agentic}} \\
\cmidrule(lr){2-5} \cmidrule(lr){6-9}
 & \textbf{Short} & \textbf{Medium} & \textbf{Long} & \textbf{Ave.} & \textbf{Short} & \textbf{Medium} & \textbf{Long} & \textbf{Ave.} \\
\midrule
\textbf{\#Samples} & 52 & 38 & 10 & 100 & 52 & 38 & 10 & 100 \\
\midrule
Qwen3-Omni-30B-A3B & 34.62 & 36.84 & 50.00 & 37.00 & 38.46 & 39.47 & 20.00 & 37.00 \\
InternVL3.5-14B    & 36.54 & 15.79 & 20.00 & 27.00 & 32.69 & 28.95 & 20.00 & 30.00 \\
MiniCPM-V 4.5      & 30.77 & 15.79 & 30.00 & 25.00 & 17.31 & 15.79 & 10.00 & 16.00 \\
Gemini-3-pro-preview & 75.00 & 65.79 & 50.00 & 69.00 & 71.15 & 84.21 & 70.00 & 76.00 \\
GPT-4o             & 46.15 & 36.84 & 40.00 & 42.00 & 51.92 & 31.58 & 40.00 & 43.00 \\
GPT-5.2            & 75.00 & 60.53 & 70.00 & 69.00 & 76.92 & 63.16 & 50.00 & 69.00 \\
\midrule
Human              & 51.92 & 50.53 & 42.00 & 50.40 & 51.92 & 50.53 & 42.00 & 50.40 \\
\bottomrule
\end{tabular}
\label{tab:v1-model-performance-duration}
\end{table*}

\subsection{Performance by Domain}

\autoref{tab:v1-model-performance-domains} reports the v1 domain-wise results. The table compares Workflow and Agentic performance across History, Geography, Culture, Economy, Technology, and Daily Life questions, and was used to analyze whether different semantic domains lead to different search-drift patterns.

\begin{table*}[t]
\centering
\small
\caption{Performance comparison across different domains under the \textbf{Workflow} and the \textbf{Agentic} settings.}
\resizebox{0.9\textwidth}{!}{
\begin{tabular}{llccccccc}
\toprule
\multirow{2}{*}{\textbf{Model}} & \multirow{2}{*}{\textbf{Setting}} & \multicolumn{7}{c}{\textbf{Domain (\%)}} \\
\cmidrule(lr){3-9}
&  & \textbf{History} & \textbf{Geography} & \textbf{Culture} & \textbf{Economy} & \textbf{Technology} & \textbf{Daily Life} & \textbf{Ave.} \\
\midrule
\multicolumn{2}{l}{\textbf{\#Samples}} & 11 & 10 & 15 & 16 & 14 & 33 & 100 \\
\midrule

\multirow{2}{*}{Qwen3-Omni-30B-A3B}
  & Workflow & 36.36 & 30.00 & 26.67 & 43.75 & 50.00 & 36.36 & 37.00 \\
  & Agentic  & 54.55 & 40.00 & 26.67 & 43.75 & 35.71 & 33.33 & 37.00 \\
\cmidrule(lr){2-9}

\multirow{2}{*}{InternVL3.5-14B}
  & Workflow &  9.09 & 50.00 & 20.00 & 25.00 & 21.43 & 30.30 & 27.00 \\
  & Agentic  & 36.36 & 40.00 & 26.67 & 31.25 & 28.57 & 24.24 & 30.00 \\
\cmidrule(lr){2-9}

\multirow{2}{*}{MiniCPM-V 4.5}
  & Workflow & 27.27 & 10.00 & 46.67 & 25.00 & 14.29 & 24.24 & 25.00 \\
  & Agentic  &  9.09 & 10.00 & 26.67 & 12.50 & 14.29 & 18.18 & 16.00 \\
\cmidrule(lr){2-9}

\multirow{2}{*}{Gemini-3-pro-preview}
  & Workflow & 72.73 & 70.00 & 80.00 & 62.50 & 64.29 & 69.70 & 69.00 \\
  & Agentic  & 81.82 & 50.00 & 86.67 & 68.75 & 85.71 & 78.79 & 76.00 \\
\cmidrule(lr){2-9}

\multirow{2}{*}{GPT-4o}
  & Workflow & 63.64 & 40.00 & 33.33 & 43.75 & 42.86 & 39.39 & 42.00 \\
  & Agentic  & 63.64 & 20.00 & 53.33 & 50.00 & 35.71 & 39.39 & 43.00 \\
\cmidrule(lr){2-9}

\multirow{2}{*}{GPT-5.2}
  & Workflow & 72.73 & 70.00 & 80.00 & 56.25 & 64.29 & 72.73 & 69.00 \\
  & Agentic  & 90.91 & 70.00 & 73.33 & 56.25 & 71.43 & 66.67 & 69.00 \\
\cmidrule(lr){2-9}

Human &  & 36.36 & 34.00 & 49.33 & 56.25 & 60.00 & 52.73 & 50.40 \\
\bottomrule
\end{tabular}
}
\label{tab:v1-model-performance-domains}
\end{table*}

\subsection{Tool-use Statistics}

\autoref{tab:v1-tool-use} reports the v1 retrieval-and-reasoning statistics, including average think calls, search calls, and runtime under both paradigms. These statistics were used to examine whether stronger performance came from more frequent tool use or from more effective evidence acquisition and filtering.

\begin{table*}[t]
\centering
\small
\setlength{\tabcolsep}{3pt}
\renewcommand{\arraystretch}{1.05}
\caption{Tool-use statistics under the \textbf{Workflow} and \textbf{Agentic} paradigms.}
\begin{tabular}{p{0.2\columnwidth} l c c c c c c}

\toprule
\textbf{Model} &
\multicolumn{3}{c}{\textbf{Workflow}} &
\multicolumn{3}{c}{\textbf{Agentic}} \\
\cmidrule(lr){2-4}
\cmidrule(lr){5-7}
& 
\textbf{\texttt{think}} & \textbf{\texttt{search}} & \textbf{time (s)} &
\textbf{\texttt{think}} & \textbf{\texttt{search}} & \textbf{time (s)} \\
\midrule

Qwen3-Omni-30B-A3B & 1.82 & 0.95 & 222.86  & 1.80 & 1.21 & 367.19 \\
InternVL3.5-14B & 1.58 & 1.48 & 214.28  & 1.20 & 1.24 & 228.05 \\
MiniCPM-V 4.5 & 0.94 & 1.13 & 242.21  & 1.97 & 2.07 & 139.07 \\
Gemini-3-pro-preview & 2.40 & 1.86 & 422.48  & 2.89 & 2.52 & 449.44 \\
GPT-4o & 1.90 & 1.11 & 136.15  & 1.31 & 1.17 & 181.87 \\
GPT-5.2 & 2.90 & 2.94 & 569.13  & 3.06 & 2.74 & 1375.83 \\
\bottomrule
\end{tabular}
\label{tab:v1-tool-use}
\end{table*}

\subsection{Error Type Distribution}

\autoref{tab:v1-error-type-distribution} reports the v1 error breakdown across categorical, incomplete, not-found, numerical, context, semantic, reasoning, and other error types. This table was used to diagnose which failure modes dominated in the previous experimental setting and how those failure modes changed between Workflow and Agentic runs.

\begin{table*}[t]
\centering
\small
\caption{Error type distribution across different models under both the \textbf{Workflow} and the \textbf{Agentic}.}
\setlength{\tabcolsep}{5pt}
\renewcommand{\arraystretch}{1.05}
\resizebox{0.95\textwidth}{!}{
\begin{tabular}{llccccccccc}
\toprule
\textbf{Model} & \textbf{Setting} &
\multicolumn{9}{c}{\textbf{Error Type Count}} \\
\cmidrule(lr){3-11}
& &
\textbf{Categorical} & \textbf{Incomplete} & \textbf{Not Found} & \textbf{Numerical} &
\textbf{Context} & \textbf{Semantic} & \textbf{Reasoning} & \textbf{Others} & \textbf{Total} \\
\midrule

\multirow{2}{*}{Qwen3-Omni-30B-A3B}
 & Workflow & 22 & 16 &  6 &  7 & 11 & 0 & 0 & 1 & 63 \\
 & Agentic  & 16 & 20 & 11 & 11 &  3 & 1 & 1 & 0 & 63 \\
\cmidrule(lr){2-11}

\multirow{2}{*}{InternVL3.5-14B}
 & Workflow & 25 & 18 & 20 &  9 &  1 & 0 & 0 & 0 & 73 \\
 & Agentic  & 29 & 17 & 15 &  7 &  0 & 1 & 1 & 0 & 70 \\
\cmidrule(lr){2-11}

\multirow{2}{*}{MiniCPM-V 4.5}
 & Workflow & 25 & 23 & 13 & 10 &  2 & 1 & 1 & 0 & 75 \\
 & Agentic  & 36 & 13 & 14 & 12 &  4 & 1 & 1 & 3 & 84 \\
\cmidrule(lr){2-11}

\multirow{2}{*}{Gemini-3-pro-preview}
 & Workflow &  5 & 10 &  4 &  9 &  0 & 2 & 0 & 1 & 31 \\
 & Agentic  &  7 &  8 &  1 &  6 &  1 & 0 & 1 & 0 & 24 \\
\cmidrule(lr){2-11}

\multirow{2}{*}{GPT-4o}
 & Workflow & 22 & 11 &  8 & 10 &  2 & 2 & 0 & 3 & 58 \\
 & Agentic  & 18 & 20 &  9 &  8 &  1 & 1 & 0 & 0 & 57 \\
\cmidrule(lr){2-11}

\multirow{2}{*}{GPT-5.2}
 & Workflow &  8 & 11 &  7 &  3 &  0 & 1 & 1 & 0 & 31 \\
 & Agentic  &  7 &  8 & 10 &  5 &  1 & 0 & 0 & 0 & 31 \\
\bottomrule
\end{tabular}
}
\label{tab:v1-error-type-distribution}
\end{table*}





\end{document}